\documentclass[10pt,twocolumn,letterpaper]{article}

 \usepackage{iccv}              % To produce the CAMERA-READY version
\definecolor{iccvblue}{rgb}{0.21,0.49,0.74}
\usepackage[pagebackref,breaklinks,colorlinks,allcolors=iccvblue]{hyperref}

\def\confName{ICCV}
\def\confYear{2025}
\def\paperID{MIRA-002}

\title{Towards a Generalizable Fusion Architecture for Multimodal Object Detection}

\author{Jad Berjawi \\
Université Grenoble Alpes, France\\
{\tt\small jad.berjawi@etu.univ-grenoble-alpes.fr}
\and
Yoann Dupas\\
Université Grenoble Alpes and Orange, France\\
{\tt\small yoann.dupas@orange.com}
\and
Christophe Cérin \\
Université Sorbonne Paris Nord and INRIA, France\\
{\tt\small christophe.cerin@univ-paris13.fr}
}

\begin{document}

\maketitle

\begin{abstract}
Multimodal object detection improves robustness in challenging conditions by leveraging complementary cues from multiple sensor modalities. We introduce Filtered Multi-Modal Cross Attention Fusion (FMCAF), a preprocessing architecture designed to enhance the fusion of RGB and infrared (IR) inputs. FMCAF combines a frequency-domain filtering block (Freq-Filter) to suppress redundant spectral features with a cross-attention-based fusion module (MCAF) to improve intermodal feature sharing. Unlike approaches tailored to specific datasets, FMCAF aims for generalizability, improving performance across different multimodal challenges without requiring dataset-specific tuning. On LLVIP (low-light pedestrian detection) and VEDAI (aerial vehicle detection), FMCAF outperforms traditional fusion (concatenation), achieving +13.9\% mAP@50 on VEDAI and +1.1\% on LLVIP. These results support the potential of FMCAF as a flexible foundation for robust multimodal fusion in future detection pipelines.
\end{abstract}    
\section{Introduction}
\label{sec:intro}

Accurate object detection is essential for reliable decision-making in real-world scenarios, where detection outcomes can directly impact safety and functionality. Consequently, reliance on visible-spectrum (RGB) images restricts the effectiveness of the model in challenging conditions, including low lighting, complex backgrounds, and occlusion.  To mitigate these limitations, recent frameworks have increasingly relied on multimodal data captured across different spectral bands, as these offer complementary and richer information. For instance, RGB images offer precise color and texture information under optimal lighting conditions, while IR captures thermal signatures that maintain reliability in poor illumination. The integration of these methodologies enables detection models to benefit from more comprehensive information. This combination is adopted in domains like autonomous driving, surveillance, and aerial monitoring  \cite{visapp25, tan2019lxmert, lu2019vilbert, bahaduri2024}.

A range of fusion strategies has been proposed in the literature to capitalize on the complementary strengths of different modalities. However, the optimal fusion strategy is dataset-dependent, as each dataset presents unique challenges that may require specific adaptation mechanisms. Different studies have investigated fusion techniques on multimodal datasets; Zhao et al. \cite{zhao2024removal} proposed fusion methods that filter noise information and then selectively choose the most relevant features. Attention-based strategies, including MEFA \cite{visapp25} and cross-channel attention mechanisms \cite{bahaduri2024}, have been developed to address these issues. However, these strategies are often tailored to specific datasets, limiting their generalization due to the implicit reliance on specific feature distributions.

In this work, we explore a generalizable preprocessing framework for multimodal fusion. This framework is designed to work across diverse conditions without requiring adaptations specific to a particular dataset. The proposed approach, Filtered Multimodal Cross Attention Fusion (FMCAF), is based on two fundamental concepts:
\begin{itemize}
    \item \textbf{Freq-Filter}: a learnable frequency-domain module that removes noisy or irrelevant information from modalities.
    \item \textbf{MCAF}: a cross-attention-based fusion module that facilitates the exchange of intermodal information, and selectively emphasizes the stronger modality per scene.
\end{itemize}

\noindent Instead of designing a pipeline specifically tailored to a dataset, the objective is to assess the generalizability of these principles. The effectiveness of the proposed approach is validated through experimentation with two distinct datasets: VEDAI \cite{vedai} and LLVIP \cite{llvip}. The experimental results demonstrate consistent improvement over the baseline traditional concatenation. These results suggest that FMCAF offers a promising and flexible starting point for robust multimodal fusion pipelines.

\section{Related Work}
\label{sec:related}
\subsection{Fusion Strategies}
The effective combination of information from multiple modalities is crucial for the success of multimodal object detection. A critical design consideration in such systems involves the stage at which fusion occurs, as this impacts the quality of the learned representations.

Fusion in multimodal systems can occur at different stages of the processing pipeline \cite{baltruvsaitis2018multimodal, ramachandra2020survey}, typically categorized as:

\begin{itemize}
    \item \textbf{Early fusion} combines raw inputs or low-level features.
    \item \textbf{Mid-level fusion} merges intermediate feature maps after modality-specific encoders.
    \item \textbf{Late fusion} combines outputs from independent networks.
\end{itemize}

In this work, a mid-level fusion strategy is adopted, whereby features from each modality are first processed independently and then integrated. However, a common approach at this stage is to concatenate modality-specific feature maps. While simple, this strategy can lead to suboptimal representations due to misalignment or conflicting noise characteristics between modalities. Thus, attention mechanisms have been introduced to guide the fusion process and enhance modality interaction.

\subsection{Attention and Cross-Attention for Fusion}
Attention mechanisms have demonstrated success in facilitating multimodal fusion by dynamically assigning relative weights to the contributions of each modality based on context. The MEFA module (Multimodal Early Fusion with Attention) \cite{visapp25} has shown strong performance in object detection by incorporating self-attention mechanisms early in the pipeline. MEFA enables the network to emphasize the most informative modality, thereby improving detection performance under varying conditions. However, MEFA relies solely on self-attention and does not support explicit cross-modal feature exchange, which limits its ability to exploit complementary cues between modalities.

On the other hand, Bahaduri et al. \cite{bahaduri2024} incorporated a cross-channel attention module that aligned RGB and IR features at an early stage in the pipeline. The proposed method involves the independent processing of each RGB channel and facilitates intermodal feature exchange through the use of a transformer-based backbone. Although these methods have proven to be effective, their implementation necessitates a substantial architectural overhead, such as tokenization, SWIN blocks, and convolutional-shifting feedforward neural networks (FFNs). Moreover, they are customized for transformer-style processing.

In contrast to MEFA, which lacks cross-modal interaction, and Bahaduri et al.'s method, which is dependent on transformer-based fusion, our approach involves an integration of cross-attention. The aim is to preserve the benefits of MEFA in terms of modality selection while facilitating early intermodal sharing.

Other cross-attention models have been proposed in the context of vision-language \cite{kim2021vilt, tan2019lxmert}, or for thermal image detection \cite{li2019illumination}, though they are typically either transformer-heavy or not compatible with real-time pipelines.

\subsection{Filtering Redundant Frequencies in Multimodal Fusion}

While the majority of multimodal fusion methods operate in the spatial domain, recent research has identified the significance of filtering redundant information in the frequency domain. Zhao et al. \cite{zhao2024removal} introduced the Redundant Spectrum Removal (RSR) module as part of a coarse-to-fine detection framework. Operating in the Fourier domain, the proposed RSR technique learns to suppress non-informative or redundant spectral components from each modality before fusion. This process reduces background clutter and enhances important features, particularly under low-light or visually noisy conditions.

Inspired by this concept, we have incorporated this frequency-domain filtering mechanism into our fusion pipeline. Specifically, a learnable spectral filtering approach is implemented for each modality before the attention-based fusion module. The objective is to enhance the quality of the joint features by reducing modality-specific noise at the input stage. In contrast to the approach proposed by Zhao et al., which integrates RSR with a downstream Dynamic Feature Selection (DFS) head, our objective is to enhance the quality of early fusion without the necessity of employing extensive post-processing modules.

By implementing frequency filtering at an earlier stage in the pipeline, we provide a more refined and informative input to the fusion module. This improves the model’s ability to focus on complementary features rather than noisy signals and contributes to improved generalization across datasets.
\section{Methods}
\label{sec:methods}
\begin{figure*}[t]
    \centering
    \includegraphics[width=1\textwidth]{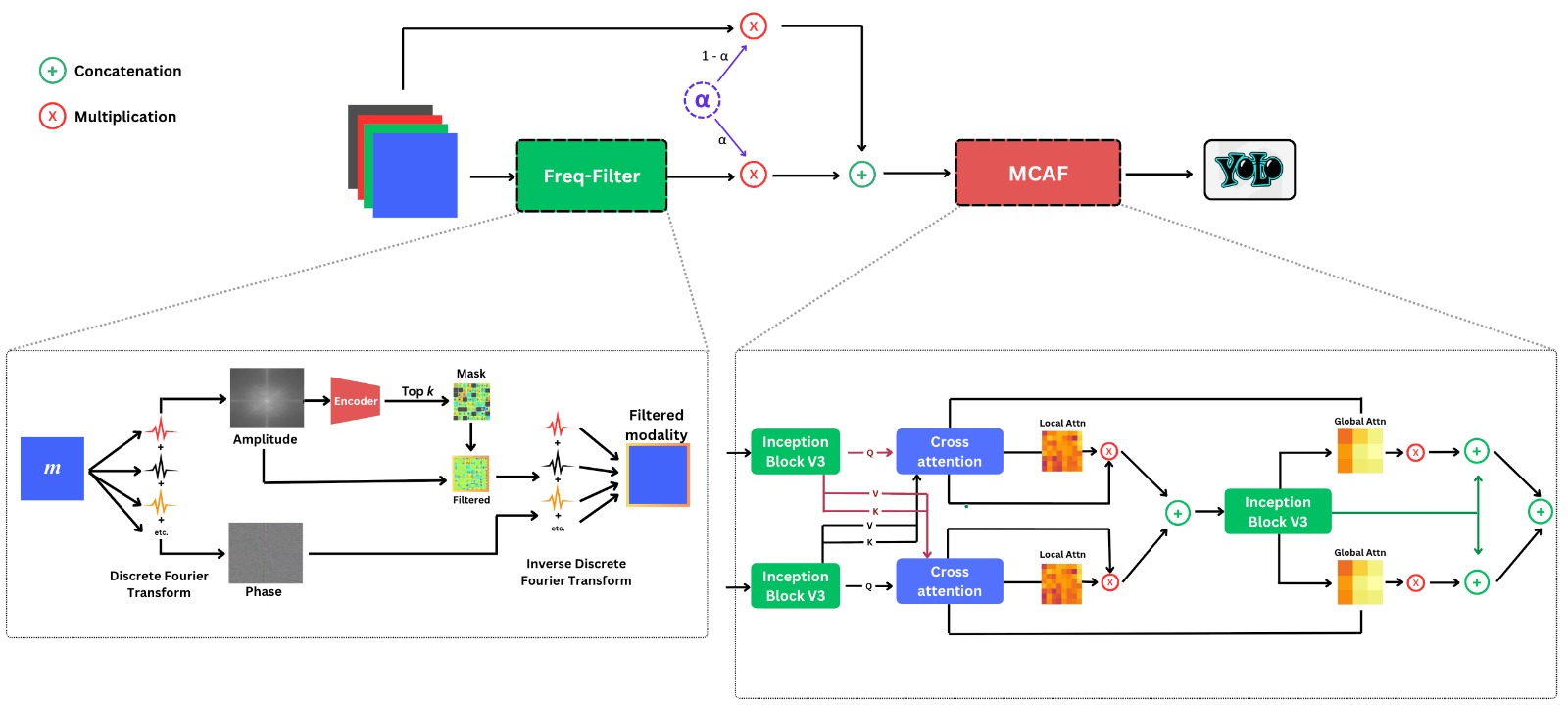}
    \caption{Overview of the proposed architecture. The Freq-Filter module (left) applies frequency-domain filtering to each modality \( m \in \{\text{RGB}, \text{IR}\} \), while the MCAF block (right) performs attention-based fusion using both self and cross-attention mechanisms.  }
    \label{fig:architecture}
\end{figure*}
\subsection{Framework Overview}
We propose a preprocessing framework designed to enhance the robustness of multimodal object detection across a range of datasets and sensor conditions. The architecture, illustrated in Figure~\ref{fig:architecture}, is based on the hypothesis that integrating frequency-domain denoising with intermodality-aware attention can mitigate noise arising from each modality and facilitate the effective exchange of complementary features across modalities.

\noindent Our method consists of two main components:
\begin{itemize}
    \item \textbf{Freq-Filter:} A frequency filter module that suppresses high-frequency noise in each modality using learnable frequency domain filters inspired by the RSR module used by Zhao et al. \cite{zhao2024removal}. This prepares cleaner modality-specific signals for fusion.
    
    \item \textbf{MCAF (Multimodal Cross Attention Fusion):} An attention-based fusion block that integrates symmetric cross-attention and hierarchical attention (local and global) to share and weigh modality features effectively.
\end{itemize}

A fundamental principle in our framework is the emphasis on flexible and learnable pre-fusion representations instead of hard fusion rules. Specifically, three architectural contributions are introduced to improve generalization and modality interaction.

\begin{enumerate}
    \item A \textbf{learnable mixing parameter} \( \alpha \in [0, 1] \) that balances raw and frequency-filtered inputs, enabling the model to adaptively control the degree of denoising during training.
    \item A \textbf{cross-attention module inserted between Inception and local attention blocks}, facilitating early sharing of complementary modality features rather than isolating them.
    \item A \textbf{residual global attention mechanism}, where global attention outputs modulate fused features through a sigmoid gate, allowing soft emphasis without erasing existing spatial cues.
\end{enumerate}

These modifications aim to make early fusion both \textit{noise-aware} (via filtering) and \textit{modality-aware} (via structured attention), while maintaining compatibility with real-time detection backbones such as YOLOv11 \cite{yolo11_ultralytics}.

Given raw RGB and IR inputs \( X \in \mathbb{R}^{H \times W \times C} \), the Freq-Filter module produces a spectrally refined version \( \widetilde{\mathbf{x}} \) by attenuating redundant frequency components. A learnable parameter \( \alpha \) is introduced to blend filtered and raw signals:

\begin{equation}
X_{\text{blend}} = \alpha \cdot \widetilde{\mathbf{x}} + (1 - \alpha) \cdot X
\label{eq:blended_input}
\end{equation}

This blended representation \( X_{\text{blend}} \) is passed to the MCAF module, which applies multi-stage attention to fuse and refine the multimodal features.

\subsection{Freq-Filter Module}

Multimodal data often includes high-frequency noise or texture artifacts that vary by modality. Inspired by the RSR module in Zhao et al. \cite{zhao2024removal}, we have integrated a frequency-domain filter to suppress these effects before fusion, thereby enabling the attention layers to focus on semantically meaningful content.

\subsubsection*{Fourier Transform and Amplitude Extraction}
Given an input tensor \( \mathbf{x} \in \mathbb{R}^{B \times 4 \times H \times W} \), composed of RGB and IR modalities, we split it into \( \mathbf{x}^{\text{RGB}} \in \mathbb{R}^{B \times 3 \times H \times W} \) and \( \mathbf{x}^{\text{IR}} \in \mathbb{R}^{B \times 1 \times H \times W} \). For each modality \( m \in \{\text{RGB}, \text{IR}\} \), we apply a 2D Fourier transform channel-wise to obtain its frequency domain representation. We then compute the average amplitude spectrum across channels to obtain a single-channel representation.

\subsubsection*{Mask Generation and Spectral Filtering}
The amplitude map \( \mathbf{A}^{m} \) is passed through a lightweight encoder to extract activations. A \emph{top-\( k \% \)} selection mechanism ranks and retains the most salient frequency components, producing a soft binary mask for each modality. This filtering reduces irrelevant frequency noise while preserving essential patterns.

The resulting soft mask is then applied to the frequency domain as it modulates the magnitude at each frequency location before the reconstruction step.

\subsubsection*{Inverse Fourier Transform}
The masked frequency signal \( \widetilde{\mathcal{F}}^{m} \) is reconstructed into the spatial domain using inverse FFT, yielding filtered output \( \widetilde{\mathbf{x}} \). Rather than hard-replacing raw input, we introduce a \textbf{learnable blending parameter} \( \alpha \), forming a weighted combination as shown in Equation \ref{eq:blended_input}. This contribution allows the model to determine during training how much filtering is useful per sample, improving flexibility and performance consistency across datasets.

While our frequency-filtering module is inspired by the Redundant Spectrum Removal (RSR) approach, we depart from the original design in how we determine the filtering threshold. In the original paper, a fixed number \(K = 320\) was selected from a total number of conceptual frequency patches. In contrast, our implementation defines a relative threshold using a ratio, \texttt{topk\_k\%}, which retains a fixed percentage of the most relevant features based on encoder output activations, making the mechanism resolution-agnostic and dynamically adaptive.

\subsection{Multimodal Cross Attention Fusion Module (MCAF)}
The MCAF module extends the MEFA attention block by integrating \textbf{cross-modal attention} and refined hierarchical attention mechanisms to support stronger intermodality feature exchange.

\subsubsection*{Cross-Attention before Local Attention}

MEFA's original design uses only self-attention, meaning each modality attends only to itself. However, this can limit the capacity to exploit cross-modal cues, especially when modalities have complementary visibility (e.g., IR for heat, RGB for texture). To address this, we insert a \textbf{cross-attention block} between the Inception-based feature extractor and the local attention stage.

Let \( m \in \{\text{RGB}, \text{IR}\} \) and \( m' \neq m \). For each modality \( X_m \), cross-attention is computed locally within non-overlapping windows of size \( w \times w \):
\begin{equation}
F_m'= \text{Softmax}\left(\frac{Q_m K_{m'}^\top}{\sqrt{d}}\right) V_{m'}
\label{eq:qkv_attention}
\end{equation}
where \( d = C/h \) is the per-head dimension and \( h \) is the number of attention heads.

This allows each modality to enrich its representation using the other's features, promoting early collaboration rather than late alignment.

% Let \( m \in \{\text{RGB}, \text{IR}\} \) denote one modality and \( m' \neq m \) its complement. Each input modality \( X_m \in \mathbb{R}^{B \times C_m \times H \times W} \) is processed through the following steps:

% We first extract low-level features using parallel Inception blocks.
% After that, we pass the output to a cross-attention block to extract inter-modal dependencies, where each modality queries the features of the other. Cross-attention is computed locally within non-overlapping windows of size \( w \times w \):

% \begin{equation}
% F_m'= \text{Softmax}\left(\frac{Q_m K_{m'}^\top}{\sqrt{d}}\right) V_{m'}
% \label{eq:qkv_attention}
% \end{equation}
% where \( d = C/h \) is the per-head dimension and \( h \) is the number of attention heads.

\subsubsection*{Local-Global Attention with Sigmoid-Gated Residual Connection}

We apply local attention to each feature map \( F_m' \) to emphasize informative spatial regions within each modality. The resulting attention maps are jointly normalized across modalities using a softmax operation. The normalized attention maps are then used to modulate the feature maps.

To produce the fused feature, we concatenate the attended modality maps and pass them through a second Inception block:
\begin{equation}
F_{\text{fused}} = \text{Inception}_{\text{fused}}\left( \text{Concat}(\tilde{F}_{\text{RGB}}, \tilde{F}_{\text{IR}}) \right)
\label{eq:inception_fusion}
\end{equation}
Following local fusion, we introduce a global attention mechanism designed to further refine the fused features. We first partition the fused feature map into non-overlapping spatial regions (8x8), and compute a global descriptor over each region. These descriptors are passed through a lightweight attention module, followed by a sigmoid activation function \( \sigma(\cdot) \):
\begin{equation}
G_m^{\text{global}} = \sigma(\text{GlobalAttn}(F_m'))
\label{eq:global_attn}
\end{equation}
Unlike softmax-based global attention, which imposes hard competition across spatial regions, the sigmoid activation allows each region to be weighted independently in the range \([0,1]\). This enables the model to simultaneously focus on multiple informative regions rather than enforcing a single dominant focus. This design choice is particularly important in scenes where multiple targets or cues are spatially distributed.

The resulting global attention map is then upsampled to match the original resolution and applied in a residual manner to preserve the original feature representation:

\begin{equation}
F_m^{\text{final}} = F_{\text{fused}} + F_{\text{fused}} \odot G_m^{\text{global}}
\label{eq:residual_fusion}
\end{equation}
This residual application connection, whereby global modulation enhances rather than replaces spatial characteristics, allows complementary signals from the local and global levels to contribute to the final fused representation.

\subsubsection*{ Final Projection to 3-Channel Output}

The final modality-specific features are concatenated and projected to produce a 3-channel fused image output, which is then passed to the object detection backbone YOLOv11 \cite{yolo11_ultralytics}.

\section{Experiments}
\label{sec:experiments}

\subsection{Datasets}

\paragraph{VEDAI (Vehicle Detection in Aerial Imagery)}
It is a publicly available dataset designed to benchmark automatic target recognition algorithms in non-constrained environments. It comprises approximately 1,200 high-resolution aerial images captured over Utah, USA. Each image is provided in both RGB and infrared (IR) modalities, with resolutions of $1024 \times 1024$ and $512 \times 512$ pixels, where only $512 \times 512$ were used for training and inference. The dataset includes annotations for 11 vehicle categories, such as cars, pickups, trucks, and camping cars. However, due to the scarcity of instances in three categories, we focus on the remaining 8 classes in our experiments, as in \cite{bahaduri2024}. VEDAI presents challenges such as small object sizes, varying orientations, occlusions, and various backgrounds, making it suitable for evaluating detection algorithms under complex conditions \cite{vedai}.

\paragraph{LLVIP (Low-Light Visible-Infrared Paired Dataset)}
 It is a dataset tailored for low-light vision tasks, including pedestrian detection, image fusion, and image-to-image translation. It contains 30,976 images, organized into 15,488 pairs of aligned RGB and thermal infrared images. These pairs are captured under various lighting conditions, predominantly in low-light or nighttime scenarios, across 24 dark and 2 daytime scenes. The images are strictly aligned in time and space, facilitating multimodal analysis. Pedestrian annotations are provided for detection tasks. For our experiments, we utilize images resized to $512 \times 512$. LLVIP presents a contrasting use case to VEDAI, focusing on \textbf{human-scale objects under degraded lighting}, which allows us to evaluate the generalization of our method across vastly different scene types \cite{llvip}.

\subsection{Evaluation Metrics}
We adopt \textbf{mAP@50 (mean Average Precision at IoU threshold 0.5)} as the principal evaluation metric. This choice reflects the benchmarks used in related multimodal detection work \cite{zhaosurvey,lin2014survey, everinghamsurvey} and is particularly appropriate in settings with alignment imprecision, small-scale targets, or diverse sensor geometries.

\subsection{Implementation Details}
All models are trained using NVIDIA A100 GPUs on the \texttt{Magi}\footnote{\url{https://github.com/Nyk0/magi-wiki}} research cluster. YOLOv11 serves as the detection backbone, and our FMCAF module is integrated as a preprocessing stage before the YOLO backbone. InceptionV3 blocks \cite{inceptionblock} are used inside the fusion module for both local and global feature aggregation.

To test the generalizability of our contributions, we train and evaluate models on both datasets separately, using 5-fold cross-validation, AdamW optimizer, an input resolution of $512 \times 512$, and dataset-specific hyperparameters.

\paragraph{VEDAI}
We train for 250 epochs to match the protocol of Bahaduri et al.~\cite{bahaduri2024}. Learning starts at 0.002 and decays to 0.01 using a cosine schedule. Momentum is 0.95, weight decay 0.01. A warm-up phase consisting of 4 epochs uses momentum 0.9, and a bias LR of 0.02. Data augmentation includes horizontal flipping (0.6), vertical flipping (0.05), rotations ($\pm$10$^\circ$), translations (10\%), scaling (30\%), HSV jittering, and heavy use of mosaic (1.0) and mixup (0.3).

\paragraph{LLVIP}
Due to the lower diversity of LLVIP, we train for only 20 epochs. The initial learning rate is 0.001, decaying to 0.01. Momentum is 0.93, weight decay is 0.001. A warm-up phase, consisting of 3 epochs, uses a lower momentum and learning rate. Augmentations include lighter rotation ($\pm$5$^\circ$), scaling (20\%), color jittering, and mixup (0.2).

Importantly, all training was conducted without dataset-specific tuning of the fusion architecture, thereby demonstrating the robustness of the proposed design under diverse sensing conditions.

\section{Results}
\label{sec:results}

\begin{table}[t]
\centering
\caption{Overall mAP@50 Improvement across datasets}
\label{tab:overall_map}
\resizebox{0.48\textwidth}{!}{
\begin{tabular}{lccc}
\toprule
\textbf{Method} & Resolution & \textbf{VEDAI (\%)} & \textbf{LLVIP (\%)}  \\
\midrule
RGB-only      & 512x512 & 62.1 & 90.2   \\
IR-only       & 512x512  & 54.2 & \textbf{97.5}  \\
\midrule
Concat (RGB+IR) & 512x512 & 62.6 & 94.3 \\
YOLOFusion \cite{YOLOFusion} & 640x640 & 73.3* &  93.1 \\
SuperYOLO \cite{SuperYOLO} & 640x640& 72.4 & 93.2 \\
\midrule
FMCAF (OURS)       & 512x512 & \textbf{76.5} & 95.4* \\
\bottomrule
\end{tabular}
}
\end{table}

\begin{figure*}[t]
    \centering
    \begin{subfigure}[t]{\textwidth}
        \centering
        \begin{minipage}[t]{0.32\textwidth}
            \centering
            \includegraphics[height=4.85cm]{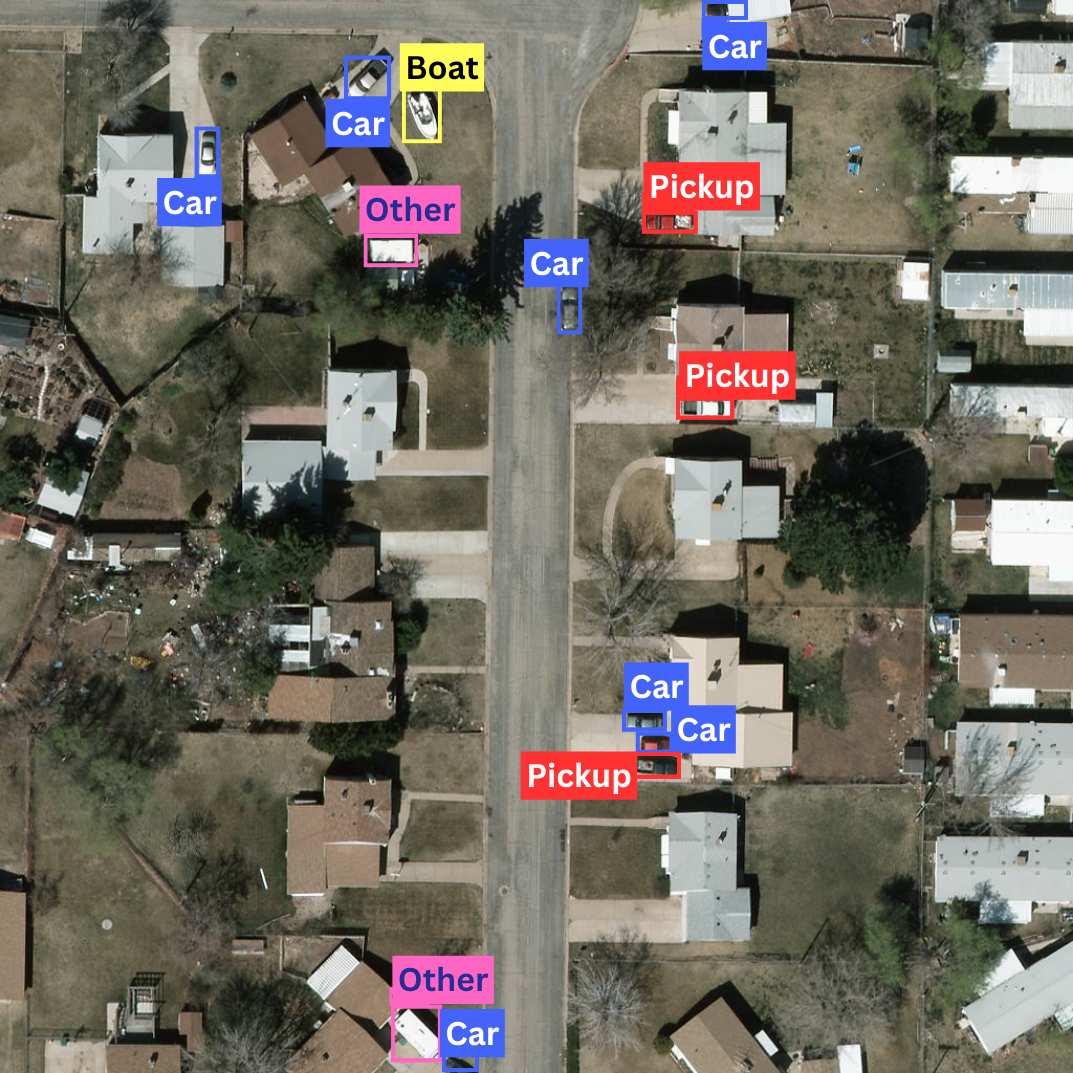}
            
            {\small GT}
        \end{minipage}%
        \hfill
        \begin{minipage}[t]{0.32\textwidth}
            \centering
            \includegraphics[height=4.85cm]{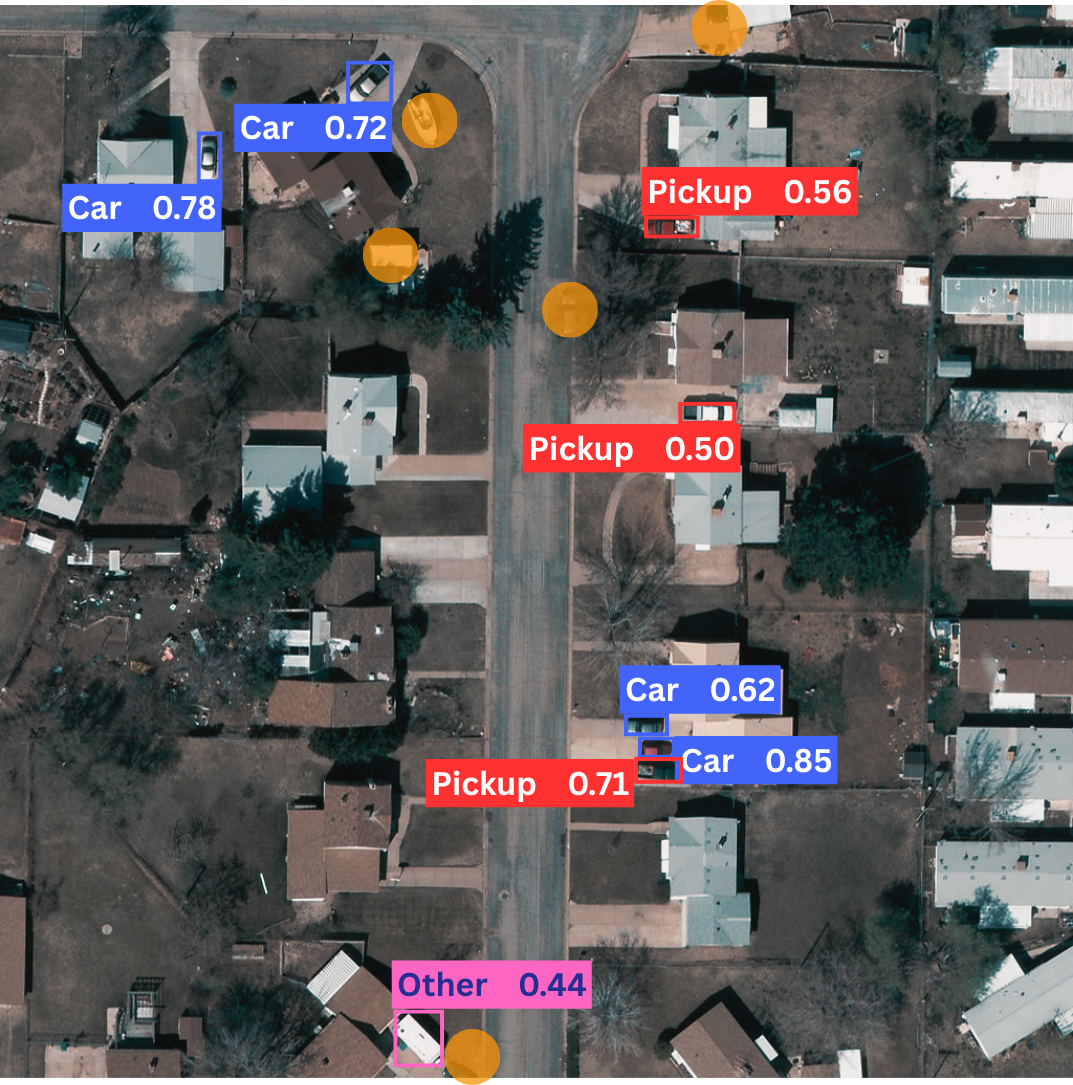}
            
            {\small Concat}
        \end{minipage}%
        \hfill
        \begin{minipage}[t]{0.32\textwidth}
            \centering
            \includegraphics[height=4.85cm]{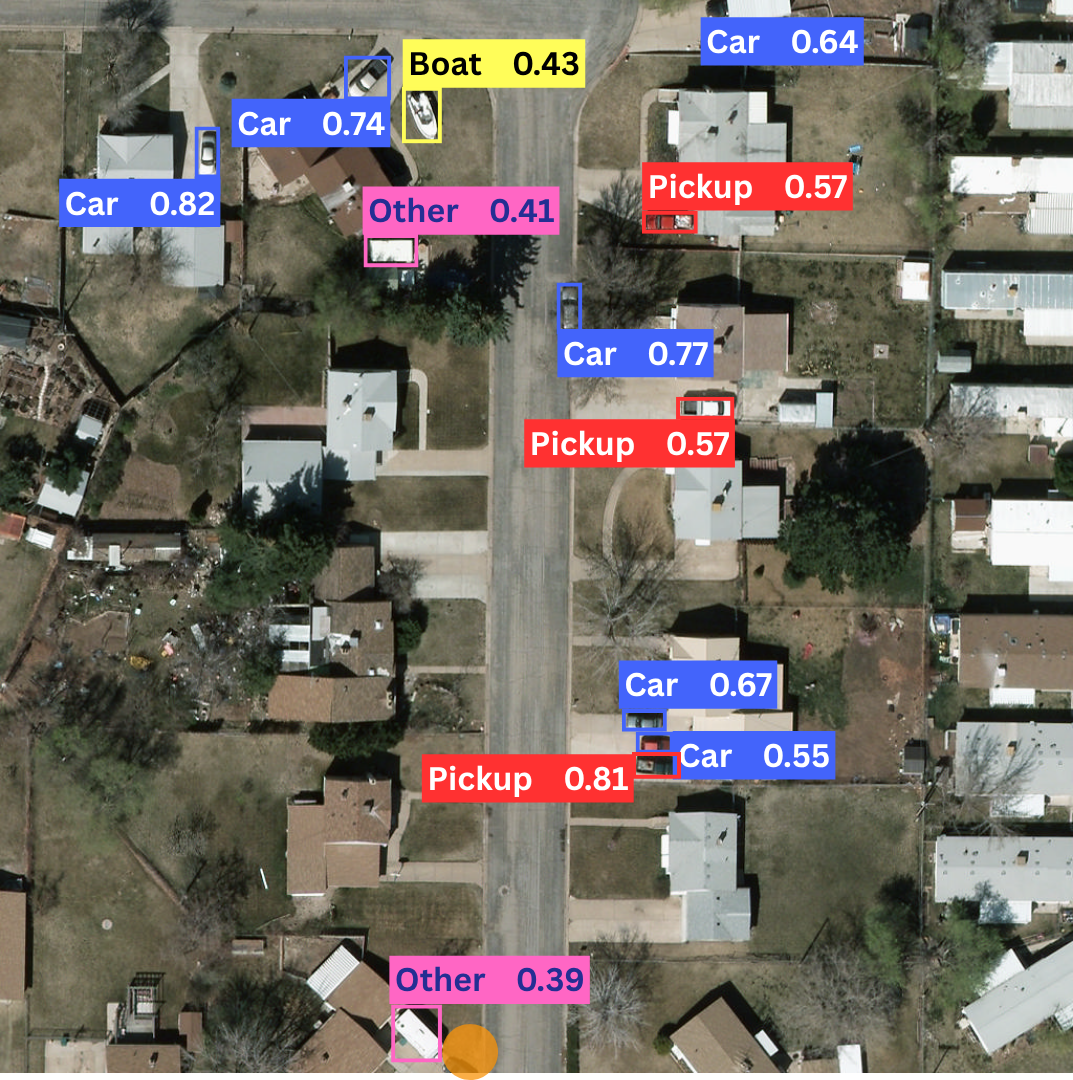}
            
            {\small FMCAF}
        \end{minipage}
        
        \caption*{\textbf{Scene 1}: VEDAI}
    \end{subfigure}
    
    \vspace{1em}
    
    \begin{subfigure}[t]{\textwidth}
        \centering
        \begin{minipage}[t]{0.32\textwidth}
            \centering
            \includegraphics[height=3.8cm]{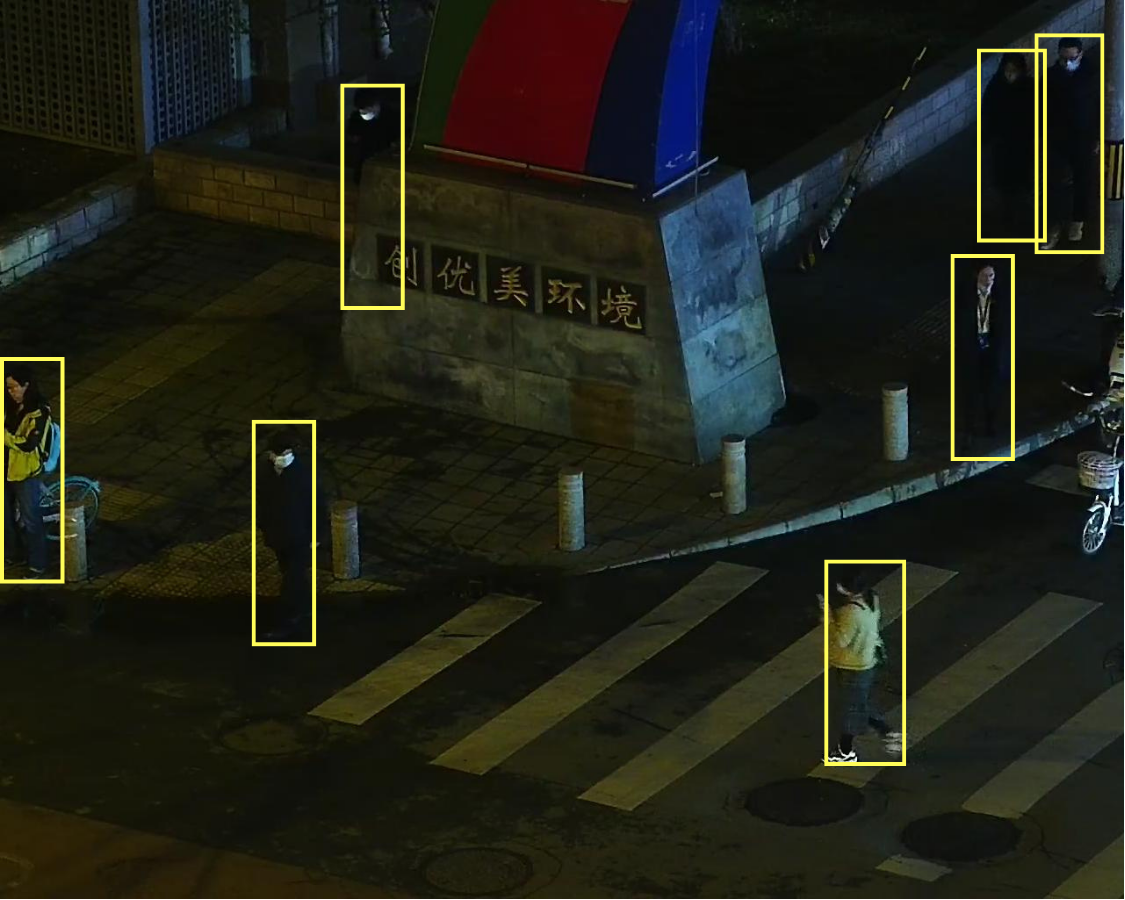}
            
            {\small GT}
        \end{minipage}%
        \hfill
        \begin{minipage}[t]{0.32\textwidth}
            \centering
            \includegraphics[height=3.8cm]{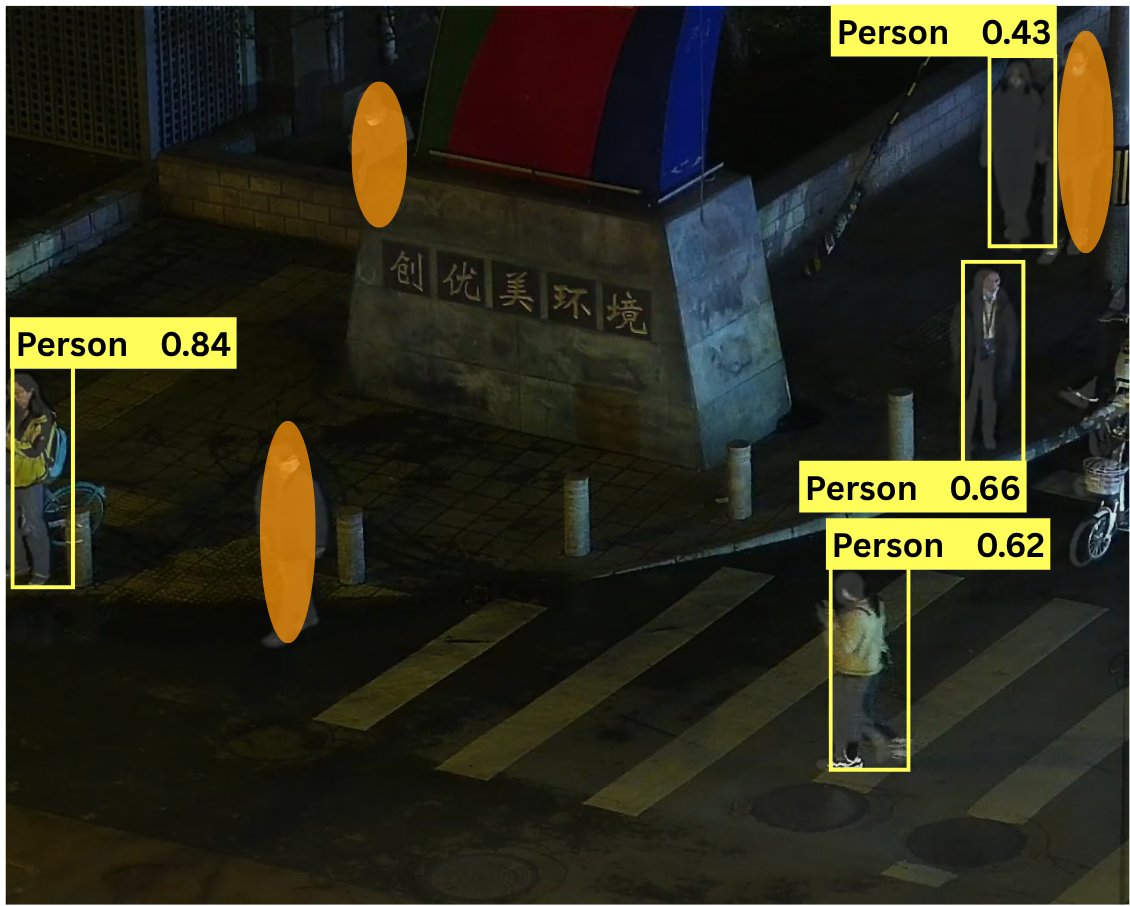}
            
            {\small Concat}
        \end{minipage}%
        \hfill
        \begin{minipage}[t]{0.32\textwidth}
            \centering
            \includegraphics[height=3.8cm]{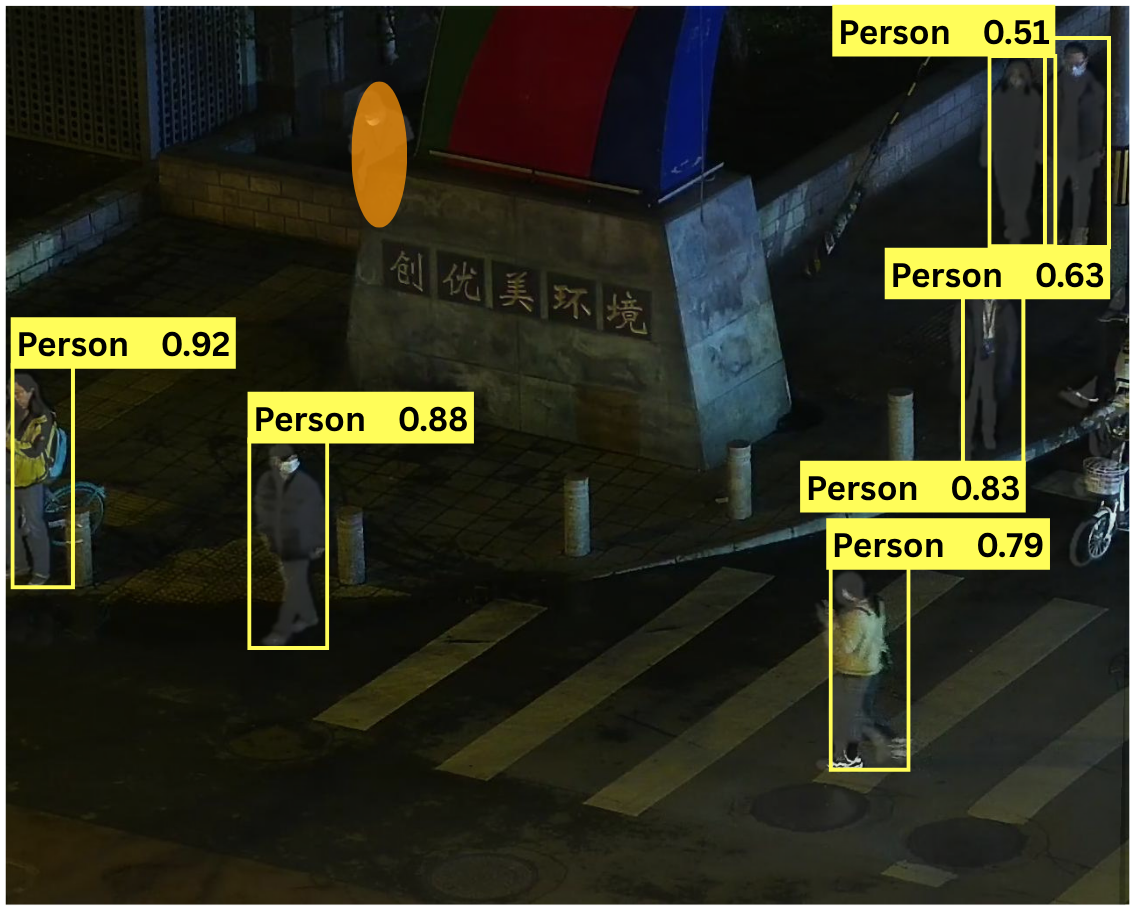}
            
            {\small FMCAF}
        \end{minipage}
        
        \caption*{\textbf{Scene 2}: LLVIP}
    \end{subfigure}
    
    \caption{Qualitative comparison of detection results across fusion methods. Each row shows one input scene, and each column presents the detection output from a different fusion configuration. Orange circles highlight missed vehicle detections.}
    \label{fig:qualitative_results}
\end{figure*}
\subsection*{Quantitative Evaluation}

We assess the performance of the proposed fusion framework by comparing four different configurations:

\begin{enumerate}
    \item \textbf{RGB Only}: The detector is trained and tested using only the RGB modality.
    \item \textbf{IR Only}: The detector is trained and tested using only the infrared modality.
    \item \textbf{Early Fusion (Concat)}: The RGB and IR channels are concatenated and used as a 4-channel input without any attention mechanism.
    \item \textbf{FMCAF}: Our full architecture combining a frequency filtering module with the attention-based module.
\end{enumerate}

Table~\ref{tab:overall_map} summarizes the mAP@50 scores for each configuration across both datasets. We report both the per-class average and the overall performance. Our proposed framework (FMCAF) achieves good accuracy in both datasets, especially by improving detection under low light or cluttered backgrounds.

Our method yields a notable improvement in detection performance, achieving an increase of +13.9\% mAP@50 on VEDAI and +1.1\% on LLVIP compared to standard early fusion by concatenation.

Although the compared fusion-based approaches (YOLOFusion \cite{YOLOFusion} and SuperYOLO \cite{SuperYOLO}) are trained and evaluated at a higher input resolution of 640$\times$640 and under slightly different initial conditions, we include them to provide a broader context for our fusion performance. This comparison, conducted under less controlled conditions, demonstrates that our FMCAF approach attains competitive accuracy even when subject to more constrained settings.

To further analyze FMCAF's strengths, we report class-wise mAP@50 results on VEDAI in Table~\ref{tab:perclass_vedai}. FMCAF achieves consistent improvements in most classes, with large performance increases observed for vehicle types with complex spatial features or limited training samples. This finding indicates that the model is particularly vulnerable to the presence of noise and that the fine-grained attention mechanism enhances feature relevance. The impact of these mechanisms is further highlighted in mid-sized classes such as {Van}, where the performance exhibits a substantial increase, rising from 56.6 (Concatenation) to 92.7 mAP@50. This point suggests that FMCAF is effective at enhancing rare class detection and can resolve modality-specific ambiguities in more common object categories.

\begin{table*}[t]
    \centering
    \caption{Performance comparison of object detection methods on the VEDAI dataset (mAP@50 per class and total).}
    \label{tab:perclass_vedai}
    \resizebox{\textwidth}{!}{
    \begin{tabular}{lcccccccccc}
        \toprule
        \textbf{Method}  & \textbf{Car} & \textbf{Pickup} & \textbf{Camper} & \textbf{Truck} & \textbf{Other} & \textbf{Tractor} & \textbf{Boat} & \textbf{Van}  & \textbf{mAP@50} \\
        \midrule
        IR-only     & 79.0 & 66.7 & 65.9 & 58.5 & 31.4 & 41.4 & 31.6 & 59.0 & 54.2 \\
        RGB-only    & 81.7 & 72.2 & 68.3 & 59.1 & 48.5 & 66.0 & 39.1 & 61.8 & 62.1 \\
        Concat (RGB+IR) & 84.3 & 72.9 & 70.1 & 61.1  & \textbf{49.9} & 67.3 & 38.7 & 56.6 & 62.6 \\
        FMCAF   & \textbf{93.6} & \textbf{84.2} & \textbf{73.7} & \textbf{93.7} & 39.6 & \textbf{72.3}  & \textbf{62.8} & \textbf{92.7} & \textbf{76.5} \\
        \bottomrule
    \end{tabular}
    }
\end{table*}
\subsection*{Inference Time and Real-Time Potential}

%We report the inference time of FMCAF as a 
We report the inference time of FMCAF as a preliminary step toward assessing its deployment feasibility. We benchmarked raw inference with a fixed input of size $X \in \mathbb{R}^{B \times 4 \times 512 \times 512}$ on an NVIDIA A100 GPU, using single-precision floating-point format (FP32). In our experiments, FMCAF achieves an average latency of 50.0 ms per image (approximately 20.0 FPS). While these values are below typical real-time thresholds (30 FPS), we emphasize that our implementation is not yet optimized for low-latency inference. This indicates that real-time deployment is feasible with hardware optimization. These results indicate that FMCAF offers a promising trade-off between detection performance and inference cost, especially considering its strong gains in multimodal robustness and adaptability.

\subsection*{Qualitative Results}

The impact of FMCAF can be presented through qualitative results shown in Figure~\ref{fig:qualitative_results}. These visualizations highlight how FMCAF performs in contrast to standard early fusion via simple concatenation.

 In the VEDAI dataset, FMCAF has been shown to identify a multitude of targets that are not recognized by the Concat baseline. This is particularly evident in instances of small or partially occluded objects, such as boats and cars. These results are consistent with the design objective of enhancing mid-level features. The enhanced spatial selectivity and elevated confidence scores indicate the model's ability to maximize the benefit of complementary modality information, surpassing the limitations of rigid concatenation strategies.

In the LLVIP scene, which presents low-light pedestrian detection challenges. The outcome of this process is detections that are both denser and more reliable, particularly in low-contrast areas where concatenation fails. These observations support our hypothesis that early attention, when supported by spectral filtering, can more effectively isolate the most informative cues, especially when one modality (IR) dominates in signal quality. 

In general, the visual output shows that FMCAF better balances the complementary properties of RGB and IR inputs, resulting in more stable detections.

% \subsection*{Ablation Study}

% To assess the contributions of each FMCAF component, we conducted ablations on the LLVIP and VEDAI datasets. Results showed that frequency filtering alone provides modest gains by reducing modality-specific noise, while cross-attention fusion (MCAF) achieves stronger improvements by enhancing spatial and semantic alignment. Combining both components yields the highest performance, validating their complementarity. We also tested different initialization values for the learnable mixing coefficient $\alpha$ and found that initializing at 0.2 leads to better convergence than extreme values (e.g., 0.01 or 0.5). Lastly, we observed that a top-k\% threshold of 75\% in the frequency filter balances noise suppression and detail retention best. Full tables and sensitivity plots are available in the supplementary material.
\section{Design Justification}

\subsection*{Performance of Frequency Filtering and Attention Fusion in Isolation}

To validate our design choices, we tested the core components independently to assess their standalone utility and complementarity:

\begin{itemize}
    \item \textbf{Freq-Filter Only}: Spectral filtering applied before standard detection backbone.
    \item \textbf{MCAF Only}: Cross-attention-based fusion without frequency filtering.
\end{itemize}

\begin{table}[h]
\centering
\caption{Component performance comparison (mAP@50).}
\label{tab:ablation_components}
\begin{tabular}{lcc}
\toprule
\textbf{Configuration} & \textbf{LLVIP} & \textbf{VEDAI} \\
\midrule
Concat (RGB+IR)         & 94.3          & 62.6           \\
Freq-Filter Only        & 91.6          & 52.5           \\
MEFA Only               & 93.5          & 70.4           \\
MCAF Only               & 94.8          & 71.8           \\
FMCAF                   & \textbf{95.4} & \textbf{74.6}  \\
\bottomrule
\end{tabular}
\end{table}

Results in table~\ref{tab:ablation_components} show that while MCAF provides strong gains on its own, especially in cluttered aerial scenes (VEDAI), frequency filtering alone struggles due to possible suppression of fine structures. The table also highlights the importance of explicit cross-modal interactions for resolving modality-specific ambiguities, as demonstrated by comparing the performance of MCAF and MEFA. Their combination in FMCAF consistently outperforms both, confirming our hypothesis that noise reduction and attention-based complementarity are synergistic, not redundant.

\subsection*{Effect of Learnable Mixing Coefficient \( \alpha \)}

The learnable mixing weight \( \alpha \) controls the balance between raw and filtered inputs. We experimented with different initialization values:

\begin{table}[h]
\centering
\caption{Impact of initial $\alpha$ value.}
\label{tab:alpha_init}
\begin{tabular}{lcc}
\toprule
\textbf{Initial $\alpha$} & \textbf{LLVIP} & \textbf{VEDAI} \\
\midrule
0.01 & 91.7         & 72.3           \\
0.2  & \textbf{95.4} & \textbf{74.6}  \\
0.5  & 93.5         & 70.4           \\
\bottomrule
\end{tabular}
\end{table}
Setting \( \alpha = 0.01 \) underutilized the filtered input, while \( \alpha = 0.5 \) risks overpowering raw spatial information. As shown in Table~\ref{tab:alpha_init}, initializing \( \alpha \) at 0.2 yields the best balance, enabling the model to learn a more effective blend.

\begin{figure}[h]
\centering
\includegraphics[width=0.48\textwidth]{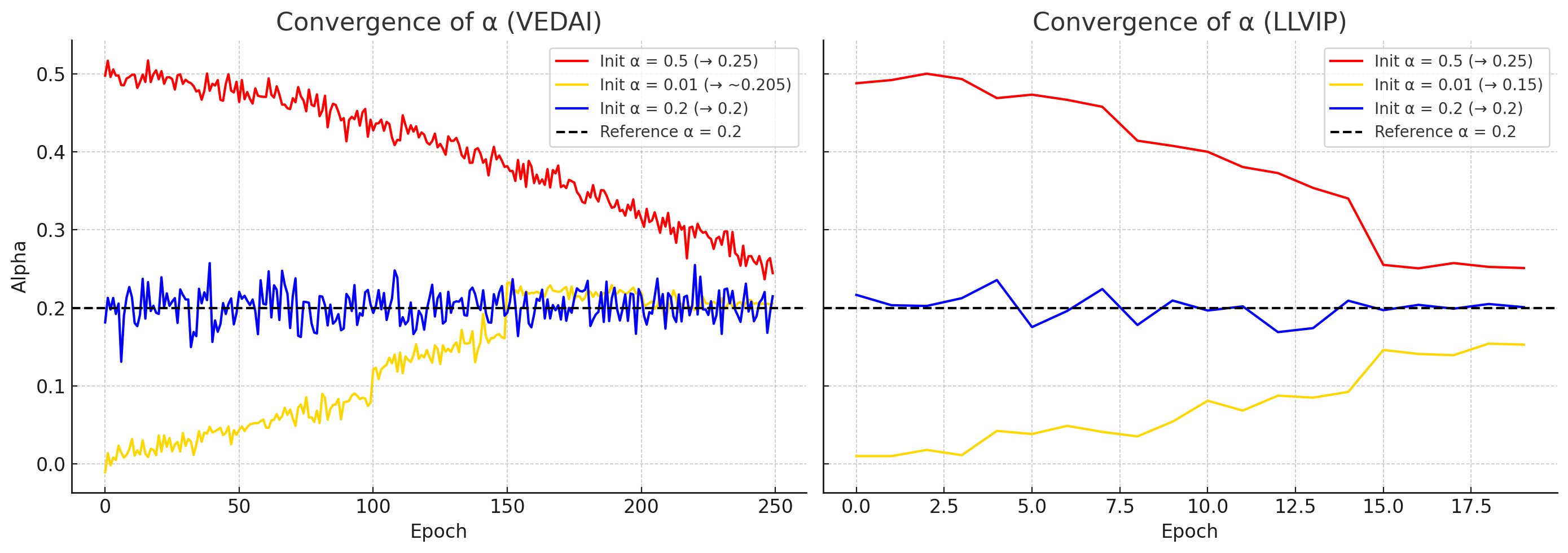}
\caption{Evolution of \( \alpha \) during training}
\label{fig:alpha}
\end{figure}

To better understand this dynamic, Figure~\ref{fig:alpha} illustrates the evolution of \( \alpha \) throughout training. The gradual adjustment indicates that the model leverages the filtered modality more as training progresses, especially once early layers have stabilized.

\subsection*{Filtering Sensitivity to Input Resolution}

Higher-resolution images are more prone to high-frequency redundancy. We investigated how input resolution affects the model’s reliance on filtered inputs by examining the learned \( \alpha \) values.

\begin{table}[h]
\centering
\caption{Average $\alpha$ at different resolutions (VEDAI).}
\label{tab:alpha_resolution}
\begin{tabular}{lc}
\toprule
\textbf{Resolution} & \textbf{Average $\alpha$} \\
\midrule
$512 \times 512$ & 0.20 \\
$896 \times 896$ & 0.32 \\
$1024 \times 1024$ & 0.45 \\
\bottomrule
\end{tabular}
\end{table}

As Table~\ref{tab:alpha_resolution} shows, the model learns to increasingly rely on frequency-filtered inputs at higher resolutions, where spatial noise becomes more prominent. This adaptive behavior reinforces the utility of our mixing mechanism.

\section{Conclusion and future works}
\label{sec:conclusion}

We introduced FMCAF, a flexible early fusion framework that jointly leverages frequency-domain filtering and cross-attention mechanisms to enhance multimodal object detection. FMCAF demonstrates strong performance across different datasets without relying on task or dataset-specific tuning by suppressing irrelevant spectral content and favoring complementary feature sharing between modalities.

Our design integrates a learnable mixing coefficient, which enables adaptive blending between raw and frequency-filtered inputs. This facilitates robustness to varying image conditions and resolution scales. FMCAF achieves a 13.9\% improvement in mAP@50 on VEDAI and a 1.1\% gain on LLVIP over standard early fusion, validating the effectiveness of combining denoising with modality-aware attention.

While current results are promising, further work is needed to explore FMCAF’s generalization beyond the evaluated datasets. Future directions include extending the architecture to additional tasks such as segmentation and classification, as well as accommodating more modalities, including unaligned or weakly calibrated sources, like cross-view medical imaging.

We also plan to investigate model compression techniques for deployment on embedded systems (e.g., ESP32, STM32), targeting edge applications such as wildlife monitoring. Preliminary post-training quantization showed limited success, suggesting that quantization-aware training may be required. An important question moving forward is how to balance model compactness with detection performance, and whether generalizable fusion architectures like FMCAF are inherently more amenable to low-power deployment than dataset-specialized counterparts.

\section*{Code and Datasets Availabilities}

Independent verification is crucial in scientific research for transparency, not just correctness. Hence, we share our Python code on an anonymous GitHub repository\footnote{\url{https://github.com/jadberjawi/FMCAF}}. The datasets used in this work are publicly available: VEDAI\footnote{\url{https://downloads.greyc.fr/vedai/}} and LLVIP\footnote{\url{https://bupt-ai-cz.github.io/LLVIP/}}.
%github link and 2 links for datasets

\end{document}